\documentclass[acmtog]{acmart}

\usepackage{booktabs} %

\usepackage[ruled]{algorithm2e} %

\SetAlFnt{\small}
\SetAlCapFnt{\small}
\SetAlCapNameFnt{\small}
\SetAlCapHSkip{0pt}

\acmJournal{TOG}
\acmVolume{39}
\acmNumber{6}
\acmArticle{256}
\acmYear{2020}
\acmMonth{12}

\setcopyright{rightsretained}
\acmDOI{10.1145/3414685.3417760}

\hypersetup{draft}

\newcommand{\el}{{\it et.~al. }}

\newcommand{\afterfigure}{\vspace{-1em}}

\begin{document}

\title{Layered Neural Rendering for Retiming People in Video}

\author{Erika Lu}
\affiliation{\institution{University of Oxford, Google Research}}
\author{Forrester Cole}
\affiliation{\institution{Google Research}}
\author{Tali Dekel}
\affiliation{\institution{Google Research}}
\author{Weidi Xie}
\affiliation{\institution{University of Oxford}}
\author{Andrew Zisserman}
\affiliation{\institution{University of Oxford}}
\author{David Salesin}
\affiliation{\institution{Google Research}}
\author{William T. Freeman}
\affiliation{\institution{Google Research}}
\author{Michael Rubinstein}
\affiliation{\institution{Google Research}}
\authorsaddresses{}

\begin{abstract}
We present a method for retiming people in an ordinary, natural video --- manipulating and editing the time in which different motions of individuals in the video occur.
We can temporally align different motions, change the speed of certain actions (speeding up/slowing down, or entirely ``freezing'' people), or ``erase'' selected people from the video altogether. We achieve these effects computationally via a dedicated learning-based layered video representation, where each frame in the video is decomposed into separate RGBA layers, representing the appearance of different people in the video. A key property of our model is that it not only disentangles the direct motions of each person in the input video, but also correlates each person \emph{automatically} with the scene changes they generate---e.g., shadows, reflections, and motion of loose clothing. The layers can be individually retimed and recombined into a new video, allowing us to achieve realistic, high-quality renderings of retiming effects for real-world videos depicting complex actions and involving multiple individuals, including dancing, trampoline jumping, or group running.

\end{abstract}

\ccsdesc[500]{Computing methodologies~Computational photography}
\ccsdesc[500]{Computing methodologies~Neural networks}

\keywords{retiming, neural rendering, video layer decomposition}

\begin{teaserfigure}
  \includegraphics[width=\textwidth]{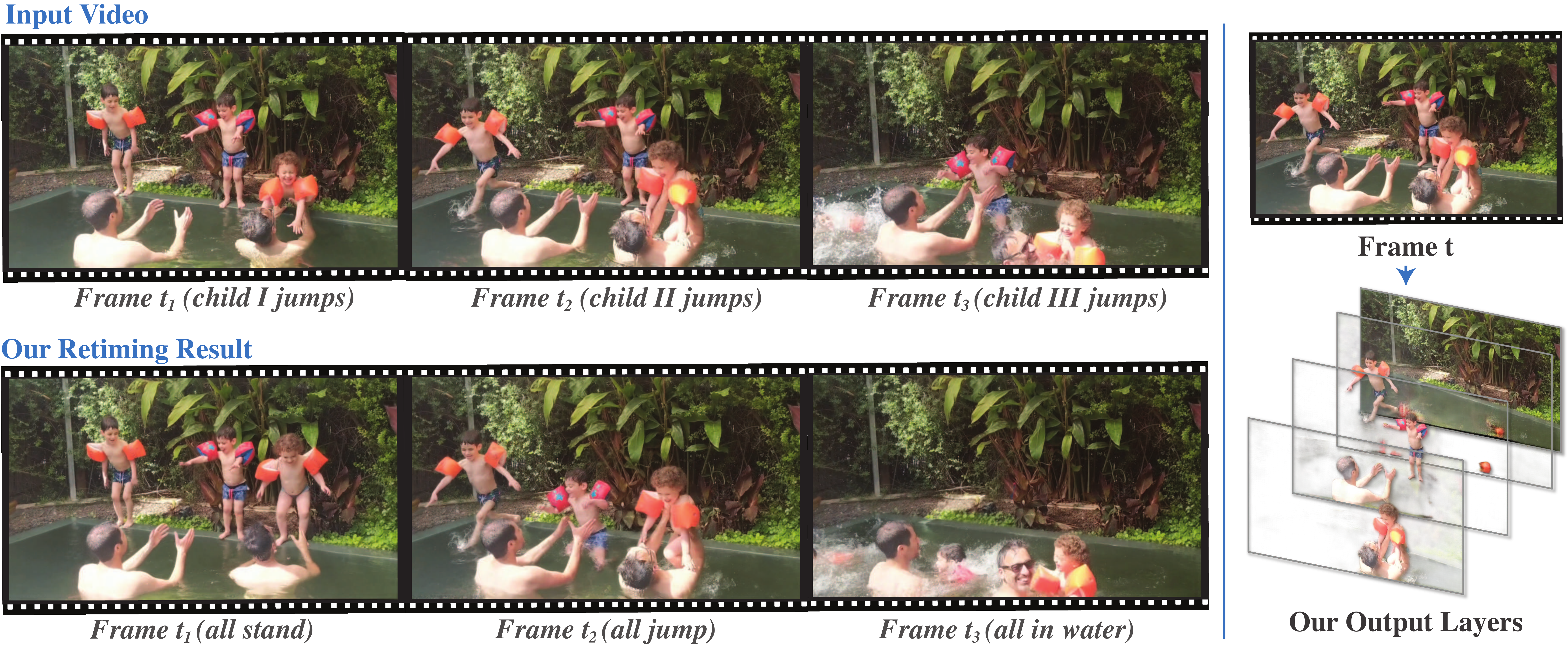}
  \vspace{-.2in}
  \caption{{\bf Making all children jump into the pool together --- in post-processing!} In the original video (left, top) each child is jumping into the pool at a different time. In our computationally retimed video (left, bottom), the jumps of children I and III are time-aligned with that of child II, such that they all jump together into the pool (notice that child II remains unchanged in the input and output frames). In this paper, we present a method to produce this and other people retiming effects in natural, ordinary videos. Our method is based on a novel deep neural network that learns a layered decomposition of the input video (right). Our model not only disentangles the motions of people in different layers, but can also capture the various scene elements that are \emph{correlated} with those people (e.g., water splashes as the children hit the water, shadows, reflections). When people are retimed, those related elements get automatically retimed with them, which allows us to create realistic and faithful re-renderings of the video for a variety of retiming effects. The full input and retimed sequences are available in the supplementary video.}
  \Description{}
  \label{fig:teaser}
\end{teaserfigure}

\maketitle

\section{Introduction}
By manipulating the timing of people's movements, we can achieve a variety of effects that can change our perception of an event recorded in a video. In films, altering time by speeding up, slowing down, or synchronizing people's motions is often used for dramatizing or de-emphasizing certain movements or events. For example, by freezing the motions of some people in an action-packed video while allowing others to move, we can focus the viewer's attention on specific people of interest. In this paper, we aim to achieve such effects computationally by retiming people in everyday videos.

The input to our method is an ordinary natural video with multiple people moving, and the output is a realistic re-rendering of the video where the timing of people's movements is modified. Our method supports various retiming effects including aligning motions of different people,  changing the speed of certain actions (e.g., speeding up/slowing down, or  entirely ``freezing'' people). In addition, our method  can also ``erase'' selected people from the video. All these effects are achieved via a novel deep neural network-based model that learns a \emph{layered decomposition of the input video}, which is the pillar of our method. Note that in this paper we focus solely on temporal warping. That is, each person's pose in our output exists in some frame in the input; we do not generate new, unseen poses or viewpoints.

Motion retiming has been studied so far mostly in the context of character animation, for editing a character's motion to match a desired duration or target velocity at a given time (e.g.,~\cite{mccann2006physics,yoo2015motion}).  In this paper, we take retiming to the realm of \emph{natural real videos}.  In the character animation domain, the main challenge is to retime the motion of a set of joints, with the spatiotemporal correlations that exist between them. Analogously, manipulating the timing of people in video not only requires modifying people's own motions, but also the various elements in the scene that they ``cause'' and are \emph{correlated} with them --- shadows, reflections, the flowing dress of a dancer, or splashing water (Fig.~\ref{fig:teaser}). When we retime people, we need to make sure that all those correlative events in the scene follow properly and respect the timing changes. Furthermore, unlike character animation, we do not have any ground truth 3D model of the scene over time; hence, rendering photorealistic, high-quality retiming effects in video is much more challenging.  More specifically, retiming motions in videos can often result in new \emph{occlusions} and \emph{disocclusions} in the scene. Rendering the scene content in disoccluded regions and maintaining correct depth ordering between subjects are essential for achieving a realistic effect. Finally, as in any video synthesis task, achieving a temporally coherent result is challenging --- small errors such as subtle misalignment between frames immediately show up as visual artifacts when the frames are viewed as a video.

The core of our technique, which we dub \emph{layered neural rendering}, is a novel deep neural-network-based model that is optimized per-video to decompose every frame into a set of layers, each consisting of an RGB color image and an opacity matte $\alpha$ (referred to altogether as ``RGBA'').   We design and train our model such that each RGBA layer over time is associated with specific people in the video (either a single person, or a group of people  predefined by the user). Crucially, our method does not require dynamic scene elements such as shadows, water splashes, and trampoline deformations to be manually annotated or explicitly represented; rather, only a rough parameterization of the \textit{people} is required, which can be obtained using existing tools with minor manual cleanup for challenging cases. The model then \textit{automatically} learns to group people with their correlated scene changes.
With the estimated layers, the original frames of the video can be reconstructed using standard back-to-front compositing. Importantly, retiming effects can be produced by simple operations on layers (removing, copying, or interpolating specific layers) without additional training or processing. 

Our model draws inspiration from recent advances in neural rendering~\cite{deferredNR}, and combines classical elements from graphics rendering with deep learning. In particular, we leverage human-specific models and represent each person in the video with a single deep-texture map that is used to render the person in each frame. Having a unified representation of a person over time allows us to produce temporally coherent results. 

We demonstrate realistic, high-quality renderings of retiming effects for real-world videos with complex motions, including people dancing, jumping, and running. We also provide insights into why and how the model works through a variety of synthetic experiments.

\begin{figure*}
    \centering
    \includegraphics[width=\textwidth]{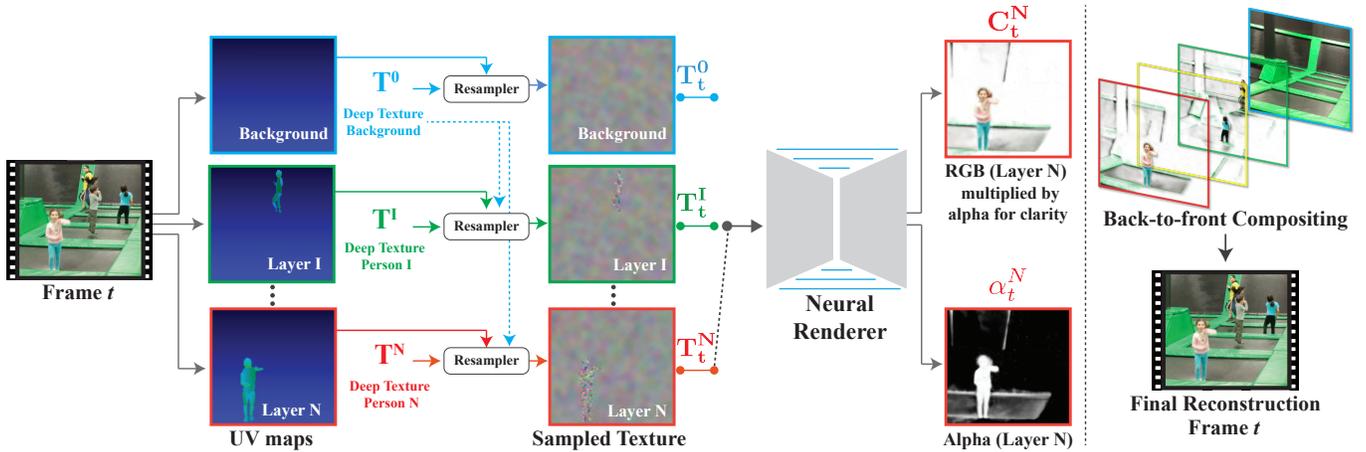}
    \caption{{\bf Layered Neural Rendering.} Our neural rendering model decomposes each frame of the video into a set of color ($C_t^i$) and opacity ($\alpha_t^i$) layers. Each  layer is associated with specific people in the video (either a single person, or a group of people predefined by the user).  The layers are computed in separate forward passes by feeding to the neural renderer a deep-texture map that corresponds to a single layer.  In particular, we represent each person in the video with a single deep-texture map $T^i$, and the scene background is represented with a deep-texture map $T^0$. Given pre-computed UV maps, those deep-texture maps are resampled and composited to form the input to the neural renderer. The set of estimated layers can then be composited in a back-to-front fashion to reconstruct the original frames. Retiming effects can be achieved via simple operations on the layers.}
    \label{fig:pipeline}
\afterfigure
\end{figure*}

\section{Related Work}

\paragraph{Video retiming.} Our technique applies time warps (either designed manually, or produced algorithmically) to people in the video, and re-renders the video to match the desired retiming. As such it is related to a large body of work in computer vision and graphics that performs temporal remapping of videos for a variety of tasks. For example, \cite{bennett2007computational} sample the frames of an input video non-uniformly to produce computational time-lapse videos with desired objectives, such as minimizing or maximizing the resemblance of consecutive frames. \cite{zhou2014time} use motion-based saliency to nonlinearly retime a video such that more ``important'' events in it occupy more time.  \cite{davis2018visual} retime a video such that the motions (visual rhythm) in the time-warped video match the beat of a target music.

Other important tasks related to video retiming are video summarization (e.g.,~\cite{lan2018ffnet}) and fast-forwarding~\cite{joshi2015real,poleg2015egosampling,silva2018weighted}, where frames are sampled from an input video to produce shorter summaries or videos with reduced camera motion or shake; or interactive manipulation of objects in a video (e.g., using tracked 2D object motion for posing different objects from a video into a still frame that never actually occurred \cite{goldman2008voa}.) 

Most of these papers retime \emph{entire video frames}, by dropping or sampling frames. In contrast, we focus on people, and people's motions, and our effect is applied at the person/layer level. While many methods exist for processing the video in the sub-frame or patch level --- both for retiming (e.g.,~\cite{pritch2008nonchronological, goldman2008voa}) and for various other video manipulation tasks, such as object removal, infinite video looping, etc. (e.g.,~\cite{agarwala05,wexler07,Barnes2010VTW,Newson14}) --- none of these works can handle automatically correlative motions and changes in the video such as shadows and reflections.

\paragraph{Manipulating human poses in video} In the context of manipulating people's motions, several recent methods have been proposed for transferring motion between two people captured in different videos~\cite{chan2019everybody, aberman2019deep, zhou2019dance, lee2020metapix} or transferring motion from a low-dimensional signal~\cite{gafni20}. However, there are two main distinctions between these  methods and ours: (i) motion transfer methods are focused on ``puppetry'' --- generating people in new unseen poses --- whereas our method is tailored solely for time warping (we do not generate new poses); and (ii) existing motion transfer methods consider only a single person in each video --- they  do not model occlusions/disocclusions between multiple people --- and with the exception of~\cite{gafni20}, they also do not capture the correlations between people and their environment, elements that are crucial for producing realistic retiming effects.

\paragraph{Image and video matting}
Our work is also related to image and video matting as we decompose each input video frame into a set of RGBA layers. Nevertheless, both traditional matting techniques (e.g.,~\cite{bai09,wang05,li05,chuang02vidmat}) as well as more recent learning-based matting methods~\cite{xu17,hou2019context} do not capture correlated effects such as shadows and reflections. Moreover, matting methods typically rely on accurate trimaps that are often given manually by the user, whereas we automatically generate rougher trimaps from detected keypoints. Even though our trimaps are not as accurate, we are still able to learn complex effects such as loose clothing, even when such regions are not included in the initial, estimated trimaps. Finally, existing matting methods cannot handle well entirely semi-transparent objects like reflections (in matting trimaps, pixels that are labeled ``opaque'' are fully assigned to the foreground). We show comparisons with image matting in Section~\ref{sec:results}.

\paragraph{Layered representations for video.}
Decomposing videos into layers is a classical problem in computer vision originally proposed in~\cite{wangadelson1994}, which has inspired a large body of work (e.g.~\cite{jojic2001sprite,zitnick04,Kumar08,Fradet08,Xue2015ObstructionFree,nandoriya2017}).
Recently, several works have leveraged the power of deep neural networks to decompose videos into layers. For example,
\cite{zhou2018stereo} and \cite{srinivasan2019} train a network on input stereo pairs of images to predict layers that correspond to physically accurate depth, with the goal of view synthesis for static scenes. Our layers are not meant to be strictly accurate in terms of depth, as we do not aim to generate scenes from novel viewpoints, but rather composite scene elements from different points of a dynamic video.

Alayrac et al.~\shortcite{alayrac2019,alayrac2019ICCV} train a model to decompose synthetically blended real videos and then apply it to natural videos to remove reflections, shadows, and smoke. Their method is general rather than person-specific like ours, and they use audio as a cue for separating layers, rather than people geometry as we do.

Finally, Gandelsman et al.~\shortcite{DoubleDIP} extend the idea of ``Deep Image Prior'' (DIP) networks \cite{deepimageprior} for decomposing a single image or a video into two layers. Their key observation is that the structure of a CNN provides a prior that drives each layer towards a natural image. As above, they have limited control over their output layers and rely entirely on CNN properties to produce meaningful layers. Our method also leverages this property of CNNs (see Section~\ref{sec:whyitworks}), yet provides control over the decomposition, which is required for retiming effects. Furthermore, their method is designed for two-layer decomposition, and generalizing it to $N$ layers requires $N\times$  learnable parameters. In contrast, our method can produce an arbitrary number of layers with the same, fixed number of network parameters. (We show a comparison with Double-DIP in the supplementary material.)

\paragraph{Neural rendering.} Neural networks have recently begun to be used as a final rendering layer for 3D scenes~\cite{lookingood:2018,meshry2019neural,neuralhumans:2019,deferredNR,sitzmann2019deepvoxels,kim2018deepvideo}. A neural network can be trained to bridge the gap between incomplete or inaccurate input geometry and a photorealistic result. Examples of incomplete input are point clouds~\cite{meshry2019neural}, voxels~\cite{sitzmann2019deepvoxels}, partial 3D scans or 3D models of humans~\cite{lookingood:2018,neuralhumans:2019,kim2018deepvideo}, and textured proxy geometry~\cite{deferredNR}.  Our work adopts the textured proxy approach --- creating geometric proxies for each person in the scene and texturing them with a deep texture map~\cite{deferredNR}. Importantly, we combine this approach in a neural rendering model that outputs a layered decomposition of each frame of the video. This is in contrast to existing neural rendering methods, which output the final reconstruction directly. Our layered decomposition is the key to producing realistic high-quality retiming effects, as demonstrated in Fig.~\ref{fig:ablation} and discussed in detail in Section~\ref{sec:ablation}.

\begin{figure*}[t!]
    \centering
    \includegraphics[width=\textwidth]{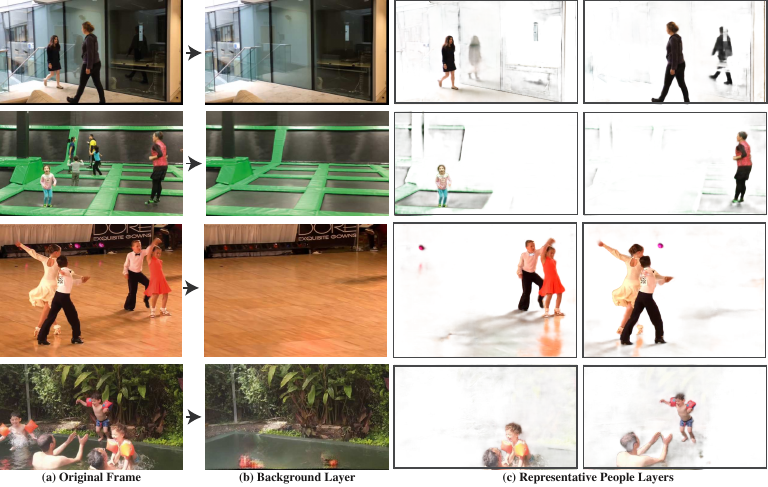}
    \caption{{\bf Example layer decompositions by our model}. For each example, we show (a) an original frame from the video, (b) the predicted background layer, and (c) two representative (RGBA) layers associated with people (in \emph{Ballroom} and \emph{Splash}, third and forth rows, respectively, the grouping of people into layers was specified manually; in the other examples each person is assigned to a separate layer). Our model successfully disentangles the people into different layers along with the visual changes they cause on the environment, such as trampoline deformations, shadows, reflections, and water splashes. (Artifacts at the bottom of the background layer in the pool scene (bottom row) are due to the fact that those regions are never visible in the input video.)}
    \label{fig:layers}
\end{figure*}

\section{Overview}
\label{sec:overview}
A key challenge in generating realistic retiming effects in videos is ensuring that when people's motions are retimed, their related effects in the scene follow with them. For example, if we freeze a person, their shadow should freeze as well.

We achieve this via a learning-based approach, which, at its core, decomposes each frame of the input video into a \emph{human-specific set of layers}. That is, each layer is associated with specific people in the video---either a single person, or a group of people (Section~\ref{sec:layers}). A key feature in our approach is that only the people in the video are modeled explicitly, while the rest of the scene elements correlated with each person are \emph{inferred automatically} by our layered neural rendering pipeline.

Our pipeline is illustrated in Fig.~\ref{fig:pipeline}. Our model is trained per-video, observing only the input video (without any additional examples), and is trained in a self-supervised manner, by learning to reconstruct the original video frames by predicting the layers (Section~\ref{sec:model}). We use off-the-shelf methods (AlphaPose~\cite{fang2017rmpe}, DensePose~\cite{guler2018densepose}) in combination with our own techniques to represent each person at each frame (Section~\ref{sec:preprocessing}). This representation is then passed on to the neural renderer and is used for seeding the layer assignment (person per layer). While the input to the neural renderer includes only the people (and a static background), the renderer's task is to generate layers that reconstruct the \emph{full} input video. Thus, it must assign all the remaining, time-varying scene elements (e.g., shadows, reflections) into the appropriate peoples' layers. 

The neural renderer succeeds in this task because the network design and training procedure (Section~\ref{sec:training}) encourages scene elements that are correlated with a layer to be captured faster (reconstructed earlier in training) than non-correlated elements. This property is related to the ``Deep Image Prior'' (DIP) work~\cite{deepimageprior}, which showed that overfitting a network to a natural image required fewer gradient descent steps than fitting a noise image. We further explore how correlated effects are captured in the correct layers through synthetic experiments in Section~\ref{sec:whyitworks}.

\section{Method}
\subsection{A Layered Video Representation}
\label{sec:layers}

Given an input video $V$, our goal is to decompose each frame $I_t\in V$  into a set of RGBA (color + opacity) layers: 
\begin{equation}
    \mathcal{L}_t= \{L_t^i\}_{i=1}^N = \{C_t^i, \alpha_t^i\}_{i=1}^N,
    \label{eq:layers}
\end{equation}
where $C_t^i$ is a color image and $\alpha_t^i$ is an opacity map (matte).  The \emph{i}$^{\textit{th}}$ layer for all frames $L_*^i$ is associated with person $i$ in the video.  We add an additional background layer $L^0_t$, not associated with any person, that learns the background color.

Given this layered representation and a back-to-front ordering for the layers, denoted by $o_t$, each frame of the video can be rendered using the standard ``over'' operator~\cite{porterduff1984}. We denote this operation by:
\begin{equation}
    \hat{I_t} =  \textit{Comp}\left(\mathcal{L}_t, o_t\right)
    \label{eq:comp}
\end{equation}

We assume that the compositing order $o_t$ is known, yet time varying, i.e., the depth ordering between people may change throughout the video (see Section~\ref{sec:preprocessing}).

A key property of this representation is that retiming effects can be achieved by simple operations on individual layers. For example, removing person $i$ from frame $t$ can be done simply by removing the \emph{i}$^{\textit{th}}$ layer from the composition, i.e., by substituting in $\mathcal{L}_t \setminus L_t^i$ into Eq.~\ref{eq:comp}. Similarly, generating a video where person $i$ is frozen at a time $t_0$ is achieved by copying $L_{t_0}^i$ over  $L_t^i$ for all frames. We expand on other operations in Section~\ref{sec:results}.

\begin{figure}[t!]
    \centering
    \includegraphics[width=\columnwidth]{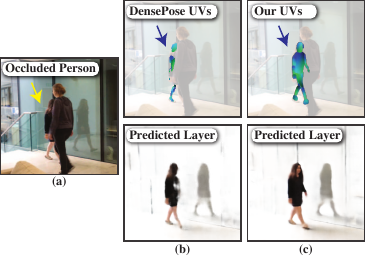} \vspace{-0.5cm}
    \caption{{\bf Full-body UVs. }Existing tools such as DensePose \cite{guler2018densepose} provide UV coordinates only for fully visible regions. In contrast, our method is able to produce full-body UVs for occluded people by detecting and interpolating keypoints, then passing the occluded person's full skeleton to a trained network that converts it to a UV map. This allows us to create editing effects that require disoccluding people.}
    \label{fig:keypoint2uv}
    \afterfigure
\end{figure}

\subsection{Layered Neural Rendering}
\label{sec:model}

Decomposing a real-world video into a desired set of layers is a difficult, under-constrained problem -- there are numerous possible decompositions that can provide an accurate reconstruction of the original frame $I_t$. For example, a single visible layer that contains the entire frame can perfectly reconstruct the video. We therefore constrain our layered neural renderer to steer the solution towards the desired person-specific decomposition.  We do so in several ways including incorporating human specific representations, tailoring the input to the network, and applying dedicated losses and training regimes.

 As mentioned in Section.~\ref{sec:overview}, the layers are predicted in \emph{separate feed-forward passes} through the neural renderer, where the input for each pass represents only one person (or a pre-defined group of people). In doing so, we control the association of people with the output layers. More specifically, we construct the input to the renderer as follows:

\paragraph{Person representation.} We parameterize each person in the video with a single human texture atlas $T^i$ and a per-frame UV-coordinate map $UV^i_t$, which maps each pixel in the human region in frame $I_t$ to the texture atlas. We use the parameterization of the SMPL model~\cite{SMPL:2015} that can be estimated from an input image using existing methods (e.g., DensePose~\cite{guler2018densepose}). This provides a unified parameterization of the person over time and a convenient model for appearance and geometry. We follow a similar approach to Thies \el \shortcite{deferredNR} and replace the classic RGB texture map with a learnable, high-dimensional texture map, which can encode more powerful and richer appearance information. The deep texture maps are then decoded into RGB values using a neural rendering network.   To represent person $i$ at time $t$, we sample its deep texture map $T^i$ using $UV_t^i$, obtaining $T_t^i$.

\paragraph{Background representation.} The background is represented with a single texture map $T^0$ for the entire video. Sampling from the background is performed according to a UV map $UV_t^0$. For a static camera, $UV_t^0$ is an identical $xy$ coordinate grid for all frames. If homography transformations estimated from camera tracking are available (see Section~\ref{sec:camera_tracking}), $UV_t^0$ is the result of transforming an $xy$ coordinate grid by the homography for frame $t$.

\paragraph{Input-Output.} To construct the input, we place the background's UV map behind each person's UV map to provide background context for the renderer (see Fig.~\ref{fig:pipeline}).  That is, the input for layer $i$ at time $t$ is the sampled deep texture map $T^i_t$, which consists of person $i$'s sampled texture placed over the sampled background texture. Each of the re-sampled texture layers $\{{T_t^i}_{i=1}^N\}$ as well as the background re-sampled texture is fed to the renderer separately. The output is  $L_t^i = \{C_t^i, \alpha_t^i\}$, the time-varying color image and opacity map for that layer, respectively. 
 
 The neural renderer is  trained to reconstruct the original frames from the predicted layers, using Eq.~\ref{eq:comp}. The depth ordering of the people layers, which may vary over time, is provided by the user. 
 To accurately reconstruct the \emph{entire} original video, the neural renderer must reconstruct any space-time scene element that is not represented by the input UV maps, such as shadows, reflections, loose clothing or hair.  Crucially, disentangling people into separate inputs is key for guiding the neural renderer to assign those scene elements into the layers with which they are most strongly correlated in space and/or time (e.g.\ attached to a person, moves like the person, etc.). We expand on this in greater detail in Section~\ref{sec:whyitworks}.

\subsection{Training}
 \label{sec:training}

We now turn to the task of learning the optimal parameters $\theta$ of the neural renderer, and the set of latent textures $\{T^i\}_{i=0}^N$ by optimizing the learned decomposition for each frame.

One necessary property of the learned decomposition is that it will allow us to accurately reconstruct the original video. Formally, 
\begin{equation}
    \mathbf{E}_\text{recon} = \frac{1}{K}\sum_t \|I_t  - Comp(\mathcal{L}_t, o_t)\|_1,
\end{equation}
where $\mathcal{L}_t$ are the output layers for frame $t$, $o_t$ is the compositing order, and $K$ is the total number of frames.

\begin{figure*}[t!]
    \vspace{-0.5cm}
    \centering
    \includegraphics[width=\textwidth]{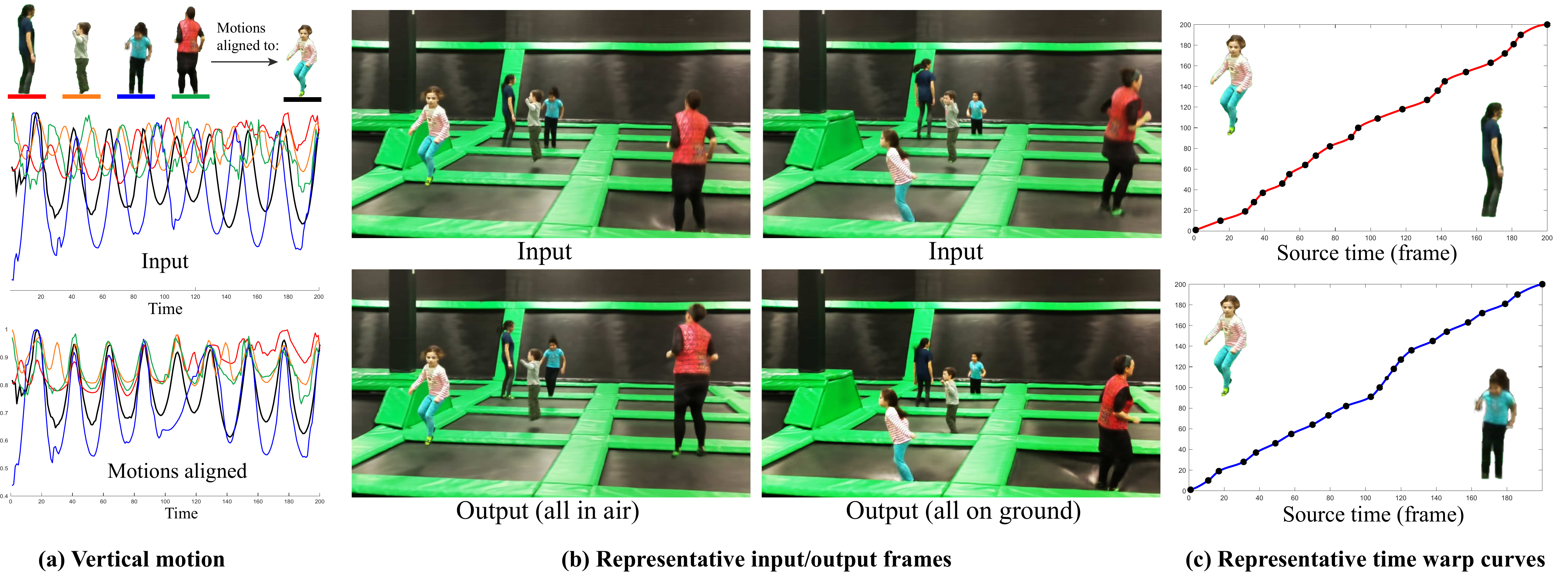} \vspace{-0.5cm}
    \caption{{\bf Automatic motion alignment}. People jumping on trampolines are motion-aligned to the little girl in the front left, such that they all jump in sync. (a) Top: motion signals for the different people in the video, taken as the normalized $y$-coordinate of the center of mass of each person over time, as computed from each person's tracked keypoints. Bottom: time-warped motion signals, after aligning each one to the motions of the little girl (black line in the plots) using Correlation Optimized Warping~\cite{tomasi2004correlation} (a variation of Dynamic Time Warping). (b) Each column shows corresponding frames from the input video (top), and from the output, retimed video (bottom). (c) Two estimated time warps, aligning two of the jumpers with the jumps of the front little girl.}%
    \label{fig:alignment}
\end{figure*}

The reconstruction loss alone is not enough to make the optimization converge from a random initialization, so we bootstrap training by encouraging the learned alpha maps $\alpha_t^i$ to match the people segments that are associated with layer $i$.  To do so, we apply the following loss:

\begin{equation}
    \mathbf{E}_{\text{mask}} = \frac{1}{K}\frac{1}{N}\sum_t \sum_i D(m_t^i, \alpha_t^i),
    \label{eq:mask}
\end{equation}
where $m_t^i$ is a trimap derived from the UV maps $UV_t^i$ (see Fig.~\ref{fig:matting}), and $D(~)$ is a distance measure. 
Masks are trimaps with values in $[0, 0.5, 1]$, where the uncertain area is produced by morphological dilation of the binary UV mask. For a trimap $m$, let $b_{0}$ be the binary mask of the pixels where $m = 0$, with $b_{1}$ defined likewise. Losses in the positive and negative regions are balanced while the uncertain area is ignored. The distance measure is:
\begin{equation}
    D(m, \alpha) = \frac{\left \| b_{1} \odot (1.0 - \alpha) \right \|_1}{2\left \| b_{1} \right \|_1} + \frac{\left \| b_{0} \odot \alpha \right \|_1}{2\left \| b_{0} \right \|_1}
\end{equation}

where $\odot$ is the element-wise (Hadamard) product.

Since the UV mask does not include information from correlated effects such as shadows and reflections, $\mathbf{E}_{\text{mask}}$ is only used to bootstrap the model and is turned off as optimization progresses.   

We further apply a regularization loss to the opacities $\alpha_t^i$ to encourage them to be spatially sparse. This loss is defined as a mix of $L_1$ and an approximate-$L_0$:
\begin{equation}
    \mathbf{E_{\text{reg}}}  =  \frac{1}{K}\frac{1}{N}\sum_t \sum_i  \gamma \left \| \alpha_t^i \right \|_1 + \Phi_0(\alpha_t^i)
        \label{eq:loss_alpha}
\end{equation}
where $\Phi_0(x) =  2\cdot \mathtt{Sigmoid}(5x) - 1$ smoothly penalizes non-zero values of the alpha map, and $\gamma$ controls the relative weight between the terms. 

Our total loss is then given by:
\begin{equation}
    \mathbf{E}_\text{total} = \mathbf{E}_\text{recon} + \gamma_m \mathbf{E}_\text{mask} + \beta \mathbf{E}_\text{reg},
    \label{eq:loss_total}
\end{equation}
where $\gamma_m$ and $\beta$ control the relative weights of the terms.

As usual with non-linear optimization, the results are sensitive to the relative weights of the error terms. To produce the results in this paper, we vary the loss weights based on the current epoch $e$ and the ``bootstrap'' epoch $e_b$, which is the first epoch where $\mathbf{E}_\text{mask} < 0.02$:

\begin{equation}
\begin{matrix}
 \gamma_{m} = \left\{\begin{matrix}
 50 & e \leq e_b \\ 
 5 &  e_b < e \leq 2e_b \\
 0 &  \text{otherwise}
\end{matrix}\right. & \gamma = \left\{\begin{matrix}
 2 & e \leq 200 \\ 
 0 &  \text{otherwise}
\end{matrix}\right. & \beta = 0.005
\end{matrix}
\end{equation}

This schedule puts a heavy initial loss on the masking term to force the optimization towards a plausible solution, then relaxes it to allow the optimization to introduce effects not present in the masks.

\subsection{High-Resolution Refinement and Detail Transfer}
\label{sec:highres}
 
We take a multi-scale approach and first train our model using Eq.~\ref{eq:loss_total} on a downsampled version of the original video. We then upsample the result to the original resolution using a separate lightweight refinement network, which consists of several residual blocks (see Appendix for details) operating on each RGBA layer separately.

We can avoid the additional expense of training with perceptual and adversarial losses by directly transferring high-resolution details from the original video in a post-processing step. The residual between the neural renderer output and the video defines the detail to transfer, and the amount of the residual to transfer to each layer is determined by the transmittance map $\tau_t^i$:
\begin{equation}
    \tau_t^i = 1.0 - \textit{Comp}_{\alpha}(\mathcal{L}_t \setminus \{ L_t^j \mid j < i \}, o_t \setminus \{ j \mid j < i\} )
\end{equation}
where $Comp_{\alpha}$ denotes the alpha channel of the composite produced by the neural renderer. The final layer colors are:
\begin{equation}
    C_t^i = Cnr_t^i + \tau_t^i (I_t  - \textit{Comp}(\mathcal{L}_t, o_t))
\end{equation}
where $Cnr$ is the color produced by the neural renderer. See supplementary material (SM) for visualizations. 
Given this transfer, the upsampling network needs only to refine the predicted alpha mattes and produce reasonable colors in occluded regions, where ground-truth high-frequency details are not available. For pixels with no occlusion and a predicted alpha value of 1, this high-frequency transfer amounts to using the original RGB pixel.

\subsection{Keypoints-to-UVs}
\label{sec:preprocessing}

Estimating people UV maps from images can be done via off-the-shelf methods (e.g., DensePose~\cite{guler2018densepose}). However, using such methods directly has two main drawbacks: 
\begin{enumerate}
    \item \emph{Disoccluded people}: 
Video retiming effects often result in disocclusions of people who were partially or even fully occluded in the original frames. Because we model the appearance of a person with a single texture map that is learned jointly for the entire video, we can render disoccluded content as long as we can correctly sample from it. Thus, we want to ensure that all UV maps represent the full body of each individual, even in the presence of occlusion. Existing methods, however, provide an estimate only for the  visible human regions (Fig.~\ref{fig:keypoint2uv}(a)). 
\item \emph{Robustness}: Existing methods tend to suffer from erratic errors such as missing body parts, especially in the presence of motion blur, noise or occlusion (see SM for examples).
Due to their dense nature, temporally filtering or interpolating UV maps in order to remove such errors is challenging. 
\end{enumerate}

 To overcome these limitation, we  train a neural network that predicts clean full-body UV maps from human keypoints. Keypoint estimators such as AlphaPose~\cite{fang2017rmpe} tend to be more robust, and unlike UV maps,  can easily be interpolated and manually corrected.   To train our keypoint-to-UV model, we use the \emph{Let's Dance Dataset}~\cite{CastroDance2017}, curated to contain only single-person video frames, and our own filmed video of approximately 10 minutes of a single person doing a variety of poses. We then generate approximately $20K$ keypoint-UV training examples in which people are fully visible, by running AlphaPose and DensePose on the original frames. We follow a similar training regimen as in \cite{guler2018densepose}; see Appendix for further details.

At test time, we can estimate full-body UV maps even in the presence of significant occlusions: we first estimate keypoints using AlphaPose, track the keypoints using PoseFlow~\cite{xiu2018poseflow}, and linearly interpolate occluded keypoints. The trained keypoint-to-UV network then processes these keypoints to generate complete UV maps (Fig.~\ref{fig:keypoint2uv}(b)). In challenging cases, current state-of-the-art keypoint detectors and trackers can still fail due to motion blur, occlusions, or noise. For the results in this paper, we manually fixed these errors using a rotoscoping tool for between zero and $10\%$ of input frames, taking a maximum of 30 minutes for particularly troublesome videos. See SM videos for examples of keypoint edits.

\subsection{Camera Tracking}
\label{sec:camera_tracking}

When the input video contains a moving camera (as is the case with all our video examples), we first estimate the camera motion using a feature-based tracking algorithm similar to the one described in~\cite{grundmann2011auto}. We model the camera motion at each frame using a homography transformation, which we estimate robustly from matched ORB features~\cite{rublee2011orb} between frames. When stabilizing small camera motions or natural hand shake, we compute the homographies between each frame and a single reference frame (which works better than tracking the camera over time), then use them to stabilize the input video. When the video contains large camera motion or substantial panning, we estimate homographies over time between consecutive frames, use them to register all the frames with respect to a common coordinate system, then apply this coordinate system transformation to the background UV map to preserve the original camera motion. Retiming a layer from frame $t$ to $\bar{t}$ is achieved by transforming the layer to the common coordinate system using the transformation at $t$, then applying the inverse transformation at $\bar{t}$.

\begin{figure}[t!]
    \vspace{-.1in}
    \centering
    \includegraphics[width=\columnwidth]{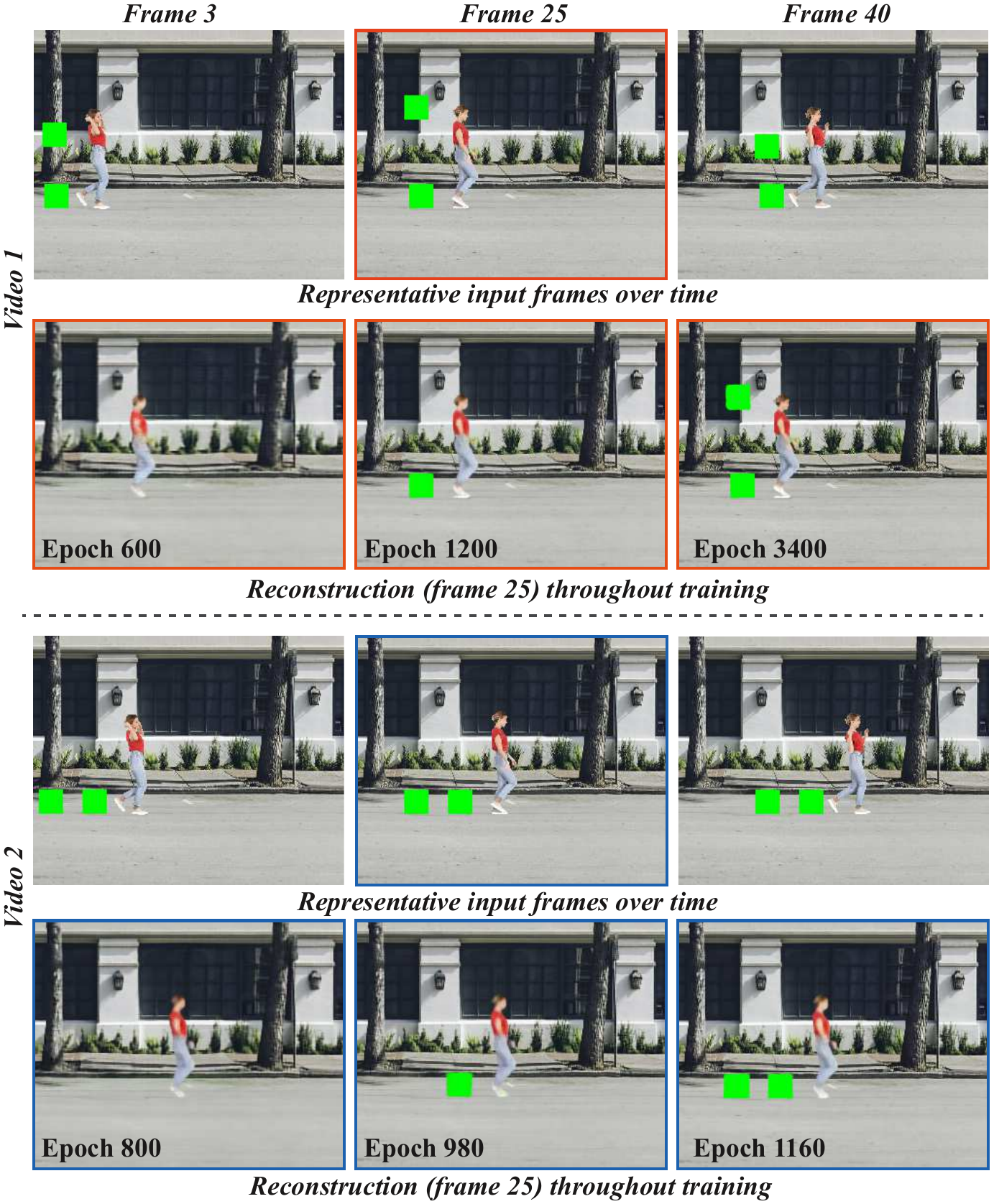}\afterfigure
    \caption{{\bf Synthetic examples for testing effects of motion correlation and spatial proximity.} In video 1, one square moves smoothly with the person while the other is randomly placed. The smoothly moving square is reconstructed earlier in training than the random square. In video 2, two squares have the same motion but one is closer to the person. The closer square is reconstructed earlier in training.}
    \label{fig:toy}
\afterfigure
\end{figure}

 \begin{figure}[t!]
    \centering
    \includegraphics[width=\columnwidth]{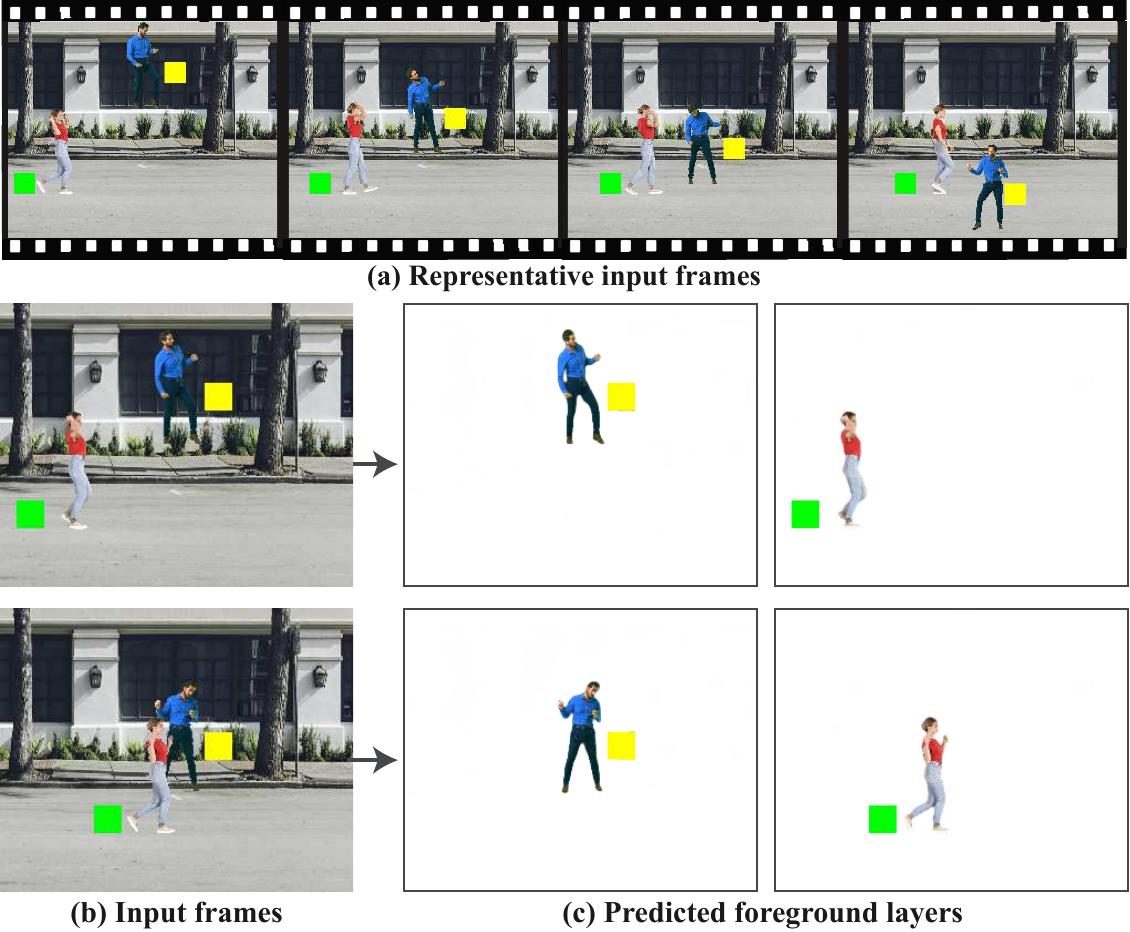}\afterfigure
    \caption{{\bf A synthetic example demonstrating assignment into layers.} (a-b) A woman (in red) is translating from left to right together with a green square next to her. Similarly, a man (in blue) is oscillating up and down with a yellow square next to him. (c) With gradient descent iterations, each person layer ``grabs'' more quickly the square that is closest and moves most similarly to it (see Fig.~\ref{fig:toy}). The squares are consistently assigned into each person's layers throughout the video, even in the presence of occlusions between the people and squares.}
    \label{fig:toy2}
\afterfigure
\end{figure}

\section{Why it works}
\label{sec:whyitworks}

As discussed in Section~\ref{sec:training}, our training ensures that specific people are assigned into specific layers (by bootstrapping the predicted alpha mattes with rough people trimaps), whereas effects outside the human trimap region, such as shadows and reflections, are learned incrementally as the network trains. But how does the model manage to associate such scene elements with the correct layers?

\vspace{-0.1cm}\paragraph{Single person and their correlated effects}
The reason for this successful assignment stems from properties of CNNs. Specifically, space-time scene elements that are most strongly correlated with the input are reconstructed earlier in training (i.e., require smaller changes to the network weights) than non-correlated elements.
To illustrate this, consider the synthetic sequences shown in Fig.~\ref{fig:toy} (the full videos are available in the SM). 
A person is superimposed on a fixed background along with two green squares. The first video examines correlations in motion and the second examines correlations in space (proximity). In the first video, one square moves consistently left-to-right behind the person, while the other is randomized around the person. The network first learns to reconstruct the static background and person, then the correlated square, and finally the random square. In the second video, two squares have the same motion, but one is closer to the person. The closer square is reconstructed earlier in training than the farther square.

This behavior is tightly related to the ``Deep Image Prior'' (DIP) principle~\cite{deepimageprior}. An image is represented by the parameters of a CNN by ``overfitting" it with gradient descent until the network output matches the image. Importantly, it was shown that fitting a natural image (which contains repetitive patches)
takes fewer iterations than fitting a random noise image (or a natural image corrupted by noise). In our case, effects that are closely correlated with a person in space, time, or both, are learned more quickly than other effects. Our synthetic examples and results on natural videos support this observation.

\begin{figure*}[htb!]
    \centering
    \includegraphics[width=.9\textwidth]{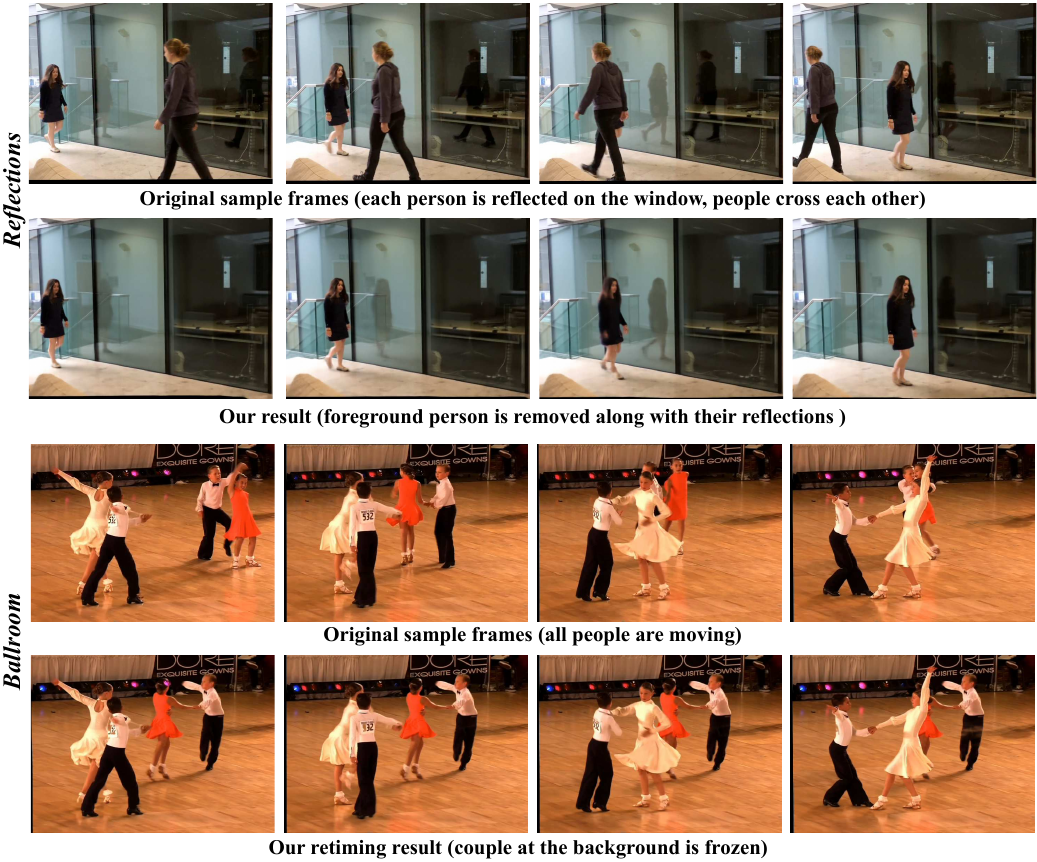} \afterfigure
    \caption{{\bf Retiming and editing results} for person removal (top) and motion freeze (bottom). For each example (pairs of rows) we show sample frames from the original video (top row) and their corresponding retimed/edited frames (bottom row).}
    \label{fig:retiming_res} \afterfigure
\end{figure*}
 
\paragraph{Assigning the correct effects to multiple people}

When there are multiple people present in the video, each person layer needs to ``grab'' the scene elements that are correlated with it. This assignment is achieved via our training regime and the construction of the input to the network. 
The renderer sees one layer's UV coordinates at a time, so must explain each missing scene element using only the information from a single person (or preselected group of people). Because correlated effects are learned faster than uncorrelated effects, 
each person layer will ``grab'' the effects that are most correlated with it before the other layers do. Once an effect is reconstructed in a layer, the reconstruction loss for that effect is minimized and no gradient encourages the effect to be learned in other layers.
In practice, this means that loose clothing, shadows, reflections, or other effects that are most strongly correlated with the person creating them, are likely to be decomposed into the correct layer.

Fig.~\ref{fig:toy2} shows such a synthetic, two-person example. The second green square from the synthetic sequence in Fig.~\ref{fig:toy} is replaced by a second person (a man in blue), who has a yellow correlated square. The motions of the two people are uncorrelated: a man (in blue) oscillates vertically while a woman (in red) translates horizontally. We bootstrap each layer with its respective person's trimap and train the network to reconstruct the video. The network learns to reconstruct the green square in the woman's layer, and the yellow square in the man's layer. Further, the yellow square (and the man) are fully reconstructed, even for frames where they were occluded by the woman or the green square.

Note that if instead of the layered input, the network sees the entire scene at one time (similar to \cite{deferredNR}), the network learns to explain the missing scene elements using \emph{combinations} of people, not individuals. Artifacts then appear when individuals are retimed to form a new combinations not seen during training (Fig.~\ref{fig:ablation}).

\begin{figure}[t!]
    \centering
    \includegraphics[width=\columnwidth]{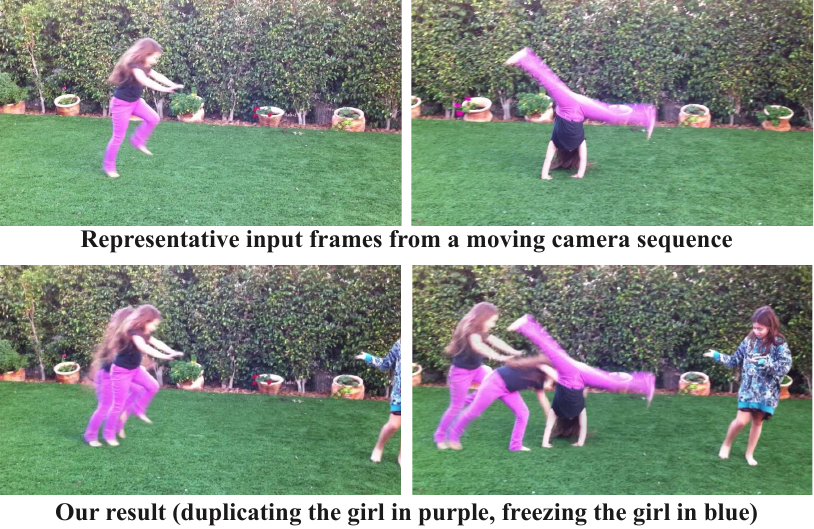} \vspace{-0.5cm}
    \caption{{\bf Cartwheel.} In the original video, the camera is panning. In our edited result, we preserve the original camera motion, while duplicating the girl who performs the cartwheel and freezing the girl in blue so that she remains in the frame rather than moving offscreen.}
    \label{fig:moving_cam}
\afterfigure
\end{figure}

\section{Results} \label{sec:results}
We tested our method on a number of real-world videos, most of which are captured by hand-held cellphone cameras (either captured by us or taken from the Web). All the videos depict multiple people  moving simultaneously and span a wide range of human actions (e.g., dancing, jumping, running) in complex natural environments. Representative frames from these videos are shown in Figs.~\ref{fig:teaser}, \ref{fig:layers} and ~\ref{fig:retiming_res}, and the full input and output sequences are available in the supplementary material.

We implemented our networks in PyTorch and used the Adam optimizer \cite{adamoptimizer} with a learning rate of $1e-3$. Depending on video length and number of predicted layers, total training time took between 5 and 12 hours on 2 NVIDIA Tesla P100 GPUs. See Appendix \ref{sec:appendix_impl} for further implementation details.

\subsection{Layer Decomposition}
Several of our layer decompositions are visualized in Fig.~\ref{fig:layers}. 
For \emph{Ballroom} and \emph{Splash}, we group certain people into one layer; for all other videos, each person is assigned with their own layer. For all the videos, our model successfully disentangles the people into the layers. The layers capture fine details such as loose hair and clothing (e.g., foreground dancer's white dress in \emph{Ballroom}), or objects attached to the people (e.g., children's floaties in \emph{Splash}).  This is in spite of initializing our model with a rough people UV map without explicitly representing these elements. This ability of our model to accurately segment the people regions is also illustrated  more closely in Fig.~\ref{fig:matting}.

Furthermore, the predicted layers successfully correlated the people with other nontrivial visual time-varying scene elements that are related to them---for example, shadows cast on the floor by the different dancers (\emph{Ballroom}), complicated reflections of two people crossing each other (\emph{Reflections}), surface deformation  (\emph{Trampoline}), or water splashes  (\emph{Splash}) caused by people's motion.

\subsection{Retiming and Editing Results}

With the predicted layers in hand, we produced a variety of retiming and editing effects via simple operations on the layers. We show several such retiming results in Fig.~\ref{fig:teaser}, Fig.~\ref{fig:alignment}, and Fig.~\ref{fig:retiming_res}  (full videos are available in the SM).

 In Fig.~\ref{fig:teaser} and Fig.~\ref{fig:alignment}, we have multiple people performing a similar action, but their motions are not in sync. In Fig.~\ref{fig:teaser}, the children jump into the pool one after the other, whereas in Fig.~\ref{fig:alignment} the periodic motions of the people jumping are independent. In both examples, we retime the people to align their motions. For \emph{Splash}, several alignment points defined manually were sufficient to align the children's jumps. In \emph{Trampoline}, because of the periodic nature of the motion, the alignment is performed automatically using  Correlation Optimized Warping~\cite{tomasi2004correlation} (a variation of Dynamic Time Warping). Notice how all the time-varying scene elements caused by the people---water splashes as they hit the water, trampoline deformations as they bounce on it---follow automatically with them in our retimed results.

 We can also use our method to ``freeze'' people at a certain point in time, while letting other people move as in the original video. This allows viewers to focus their attention on the moving people while ignoring the rest of the motions in the original video.  In \emph{Ballroom} (Fig.~\ref{fig:retiming_res}), we freeze the dancing couple in the back throughout  the video, while the couple in front keeps moving. Here too, the shadows and reflections on the floor move realistically with the moving couple, while the shadows of the background couple remain static. Furthermore, disoccluded regions of the back couple are realistically rendered.

In \emph{Kids Running} (supplementary material), 
we show how our model can scale to multiple layers (8) to produce complex retiming effects involving many people. We retime the original video, where the kids are crossing the faint finish line on the ground at different times, to produce a `photo-finish' video, where all the kids cross the finish line together. We achieve this result by slowing down the layers of the children that run offscreen. Even though this sequence involves many individuals, our model is able to obtain clean mattes for each child. In addition,   occluded people and large occluded regions in the background are realistically inpainted, all while handling significant motion blur that exists in the input video.

As mentioned, in addition to retiming effects, our method can also support easy removal of people in video---a byproduct of the layered representation we use. In \emph{Reflections}, we demonstrate person removal in a video containing two people crossing paths in front of a window. Here our model manages to perform several nontrivial tasks: it completely disoccludes the person walking in the back; it associates each person properly with their reflection and shadow; and it disentangles the two reflections when they overlap (despite the fact that none of these elements are represented explicitly by the model). Consider generating such a result with a traditional video editing pipeline: the reflections would have to be tracked along with the people to perform proper removal; the person in the back would have to be manually inpainted at the point where they are occluded by the person in front. The advantage of our method is that by merely inputting UVs for each person in a separate layer, and turning those layers `on and off', we can achieve the same result with significantly less manual work.

In Fig.~\ref{fig:moving_cam}, we show an example where the original camera motion is preserved in our retimed result (see Section~\ref{sec:camera_tracking}) -- the girl on the left is duplicated with a short time offset between each of her copies, and the girl on the right is frozen, all while the camera is panning as in the original video.

\subsection{Comparisons}
\begin{figure*}[t!]
    \centering
    \includegraphics[width=\textwidth]{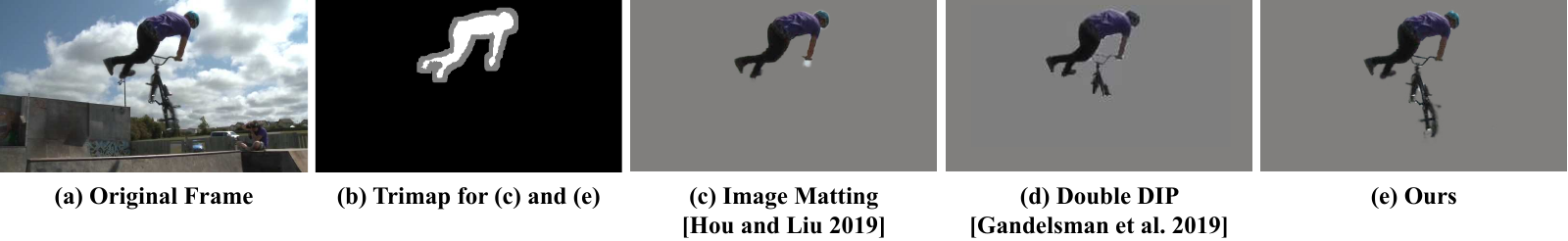}\afterfigure
    \caption{{\bf{Comparison with matting and Double-DIP.}} We obtain trimaps from a mask of the person's UV; the gray pixels represent a dilated region around this mask for which the matching loss is ignored. Matting is insufficient for our retiming task because it fails to capture correlated effects that are not included in the input trimap (e.g. the bicycle). Unlike matting, our final model prediction is able to move beyond the bounds provided by the coarse trimap to produce a more accurate segmentation mask that includes correlated regions. While Double-DIP is not limited by trimap bounds, it fails to segment the entire bicycle.}
    \label{fig:matting}
    \afterfigure
\end{figure*}

We compare our method to state-of-the-art learning-based image matting~\cite{hou2019context} and Double-DIP~\cite{gafni20} in Fig.~\ref{fig:matting}. Matting methods are unsuitable for retiming because they fail to capture desired effects outside of the trimap region. To capture such effects using matting would require annotating the trimap to represent these regions. Our method, however, is not strictly bound by the trimap, as we use it to bootstrap our training only in the initial epochs, after which the optimization is allowed to grow the matte to encompass regions beyond the trimap border. Thus, we can capture complex effects \textit{without explicitly representing them in the trimap}, merely by representing the human region coarsely using automatically generated trimaps. While Double-DIP is similarly unbound by the trimap region, it fails to segment the entire bicycle.

 \begin{figure}[t!]
    \centering
    \includegraphics[width=.42\textwidth]{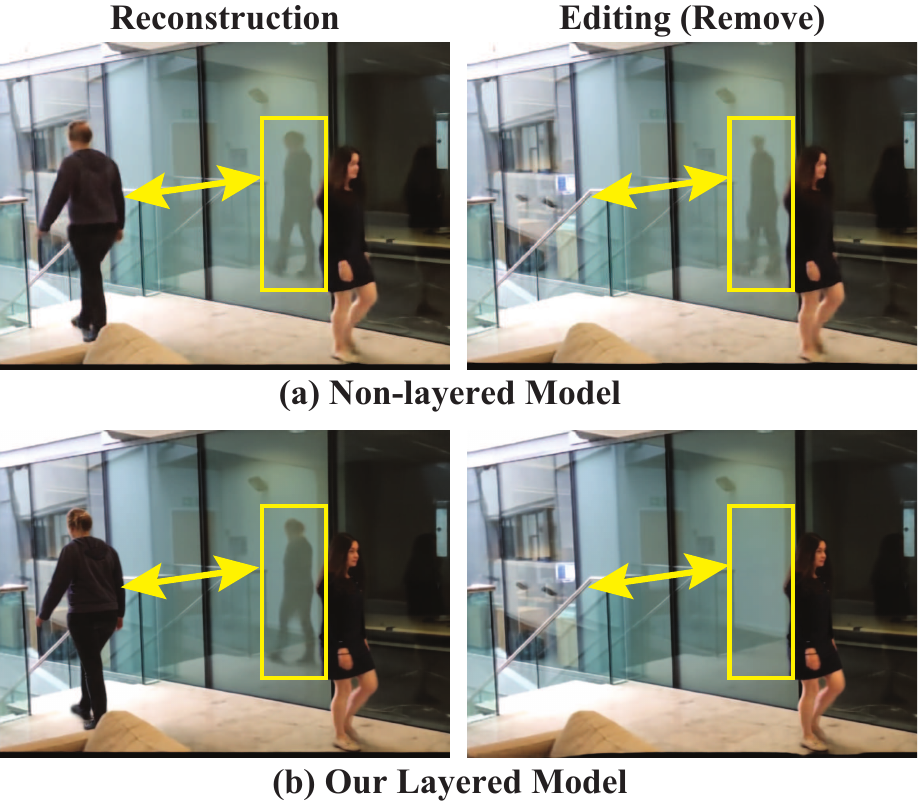}\afterfigure
    \caption{{\bf Layered vs. non-layered neural rendering.} Removing the layer decomposition component of our model results in a lack of generalization ability. While both models are able to reconstruct the original video, the non-layered model struggles to disentangle people and their correlations (reflection in right column is not fully removed when the person is removed).} \afterfigure
    \label{fig:ablation}
\end{figure}

\subsection{Ablations}
We ablate components of our method, specifically the keypoint-to-UV network and layer decomposition.
\label{sec:ablation}
\subsubsection{Keypoint-to-UV network vs. DensePose}
We compare layer decomposition results from using our keypoint-to-UV network to predict UV maps versus using DensePose outputs directly (Fig.~\ref{fig:keypoint2uv}). In the case of partial occlusion, our UV maps manage to complete the occluded regions of the person, while the DensePose UVs fail to complete the face and the left arm. In a more extreme case where the person is fully occluded, but their reflection is visible, it is crucial to generate a UV map for that person not only for the purposes of disocclusion, but because any visible effects such as their reflection or shadows will be incorrectly reconstructed in the layer of the visible person (as the fully occluded UV will be indistinguishable from the background-only UV). Thus we disocclude UVs by using keypoints as an intermediate representation that allows for more robust estimation and straightforward interpolation.

\subsubsection{Layered vs. Non-Layered Neural Rendering}
We compare our method to a non-layered model, i.e., a model that takes as input a single sampled texture map representing all the people in the frame as well as the background, and directly outputs an RGB reconstruction of the frame.  This baseline follows the deferred neural rendering approach~\cite{deferredNR} . 

Fig.~\ref{fig:ablation} shows a reconstruction and editing result produced by the non-layered model. As can be seen, the non-layered model can reconstruct the original frames fairly well despite the missing information and noise in the UV maps. This aligns with the results demonstrated in \cite{deferredNR}. However, when editing is performed, the non-layered model performs poorly and is unable to generalize to new compositions of people, as evident by the partial reflection artifact. The main reason is twofold: (i) To produce editing effects, the model is required to generalize to new UV compositions of people in configurations that were never seen during training; thus it struggles to produce realistic-looking results based solely on L1 reconstruction loss. Our approach avoids this generalization issue because editing is performed as post-processing on the predicted layers---the same outputs produced during training.
(ii) When the input to the model is a composition of all the people in the frame rather than separated UVs, the model can easily reconstruct the original frame without necessarily capturing meaningful correlations since it is not required to disentangle separately moving parts of the scene. This can be seen where the non-layered model struggles to learn the correct relationships between the different people and their reflections.

 Another benefit of the layered representation is that we can transfer high-resolution details from the input video to each layer, as described in Section \ref{sec:highres}. Because the non-layered model produces the retimed result directly, there is no opportunity to transfer detail from the input video prior to retiming. Achieving comparable quality with the non-layered model would likely require extensive training time and additional loss functions.

 \begin{figure}[t!]
    \centering
    \includegraphics[width=0.47\textwidth]{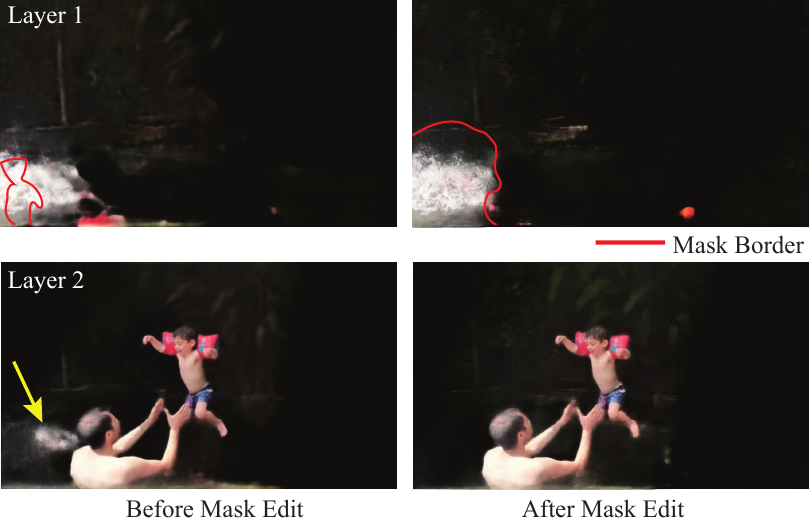} \afterfigure
    \caption{{\bf{User-guided trimaps.}} In some particularly difficult segments, the assignment of effects to layers may be incorrect, e.g., some parts of the splash caused by the child who is underwater are incorrectly assigned to Layer 2  (yellow arrow). The user can correct such errors by simple manual editing of the initial trimaps. Here, we expand the initial trimap to include the entire splash region (red outline); this was done for only one layer of one frame, and then duplicated for a segment of the video.}
    \label{fig:fix_masks}
\afterfigure    
\end{figure}

\subsection{Limitations}

While our system manages to decompose scenes quite well---even surprisingly so---in some scenes the results may not be perfect. Here we point out some of the limitations that we have observed.

For a particularly difficult section of the \textit{Splash} video, part of the large water splash caused by the child who is underwater in \emph{Layer 1} is incorrectly placed in \emph{Layer 2} (Fig.~\ref{fig:fix_masks}). However, this error can be fixed by manually editing the initial trimap masks for Layer $1$ to include the entire splash (top right, Fig.~\ref{fig:fix_masks}). Such editing can be done very quickly when needed, requiring only rough marking on a few of the frames (a single trimap in this example).

Another type of artifact that may occur is when the scene includes time-varying background elements that are not correlated with any of the people of interest in the video. For example, since we assume a static background, the colorful blinking lights in the \emph{Ballroom} video must be included in one of the foreground layers. As can be seen in Fig.~\ref{fig:layers}, the purple light appears in the front couple's layer, making it impossible to retime them separately. This artifact could be remedied by adding a separate layer to represent the dynamic background elements in addition to the layers for the dancing couples.

\section{Conclusion}

We have presented a system for retiming people in video, and demonstrated various effects, including speeding up or slowing down different individuals' motions, or removing or freezing the motion of a single individual in the scene. The core of our technique is learned layer decomposition in which each layer represents the {\em full} appearance of an individual in the video --- not just the person itself but also all space-time visual effects correlated with them, including the movement of the individual's clothing, and even challenging semi-transparent effects such as shadows and reflections.

We believe that our layered neural rendering approach holds great promise for additional types of synthesis techniques, and we plan to also generalize it to other objects besides people, and to expand it to other non-trivial post-processing effects, such as stylized rendering of different video components.

\section*{Acknowledgments}
We thank the friends and family that appeared in our videos. The original \emph{Ballroom} video belongs to Desert Classic Dance. This work was funded in part by the EPSRC Programme Grant Seebibyte EP/M013774/1.

\bibliographystyle{ACM-Reference-Format}
\bibliography{refs}

\appendix
\section{Appendix}
\label{sec:appendix}

\subsection{Implementation Details}
\label{sec:appendix_impl}
\paragraph{Neural texture.} We use a neural texture with 16 channels, where each body part and the background are represented by an atlas with $16\times 16$ pixel dimensions. As in the DensePose work, we adopt the SMPL representation, which uses 24 body parts. Thus the dimensions of our neural texture, for a video with $N$ people, is $16\times16\times16*(24N+1)$. Empirically we found that increasing the texture resolution did not improve results.
\paragraph{Training on low-resolution.} For each video, we first train the neural renderer and neural texture for $2500$ epochs at a video size of $448\times256$ (or $352\times 256$ for the \emph{Ballroom} sequence only). The input to the neural rendering network is the sampled neural texture concatenated with a map representing the person ID at each pixel (or 0 for background). We apply brightness and spatial jittering to the video frames as data augmentation. Brightness jittering is turned off after 400 epochs to allow the network to faithfully reconstruct the original video. We also employ curriculum learning, where we train only on the easier half of the frames for the first 1k epochs. We rank difficulty by computing the IoU of every pair of layers' positive trimap regions, using distance of bounding box centers as a tie-breaker. The goal of this curriculum is to withhold the more difficult frames that contain overlapping people, because of the greater ambiguity in learning correlations.
\paragraph{Upsampling low-resolution results.} After training the neural rendering module, we obtain a high-resolution result by freezing the trained parameters and training an additional upsampling network. This lightweight upsampling network is trained for 500 epochs with only L1 reconstruction loss, sampling random $256\times256$ crops for upsampling due to memory constraints. The final output of the upsampling network has dimensions double the size of the low-resolution output. We apply the detail transfer step described in Section \ref{sec:highres} to these final outputs.

\subsection{Network Architectures }
In all networks, zero-padding is used to preserve the input shape. `bn' refers to batch normalization. `in' refers to instance normalization. `convt' refers to convolutional transpose. `leaky' refers to leaky relu with slope -0.2. `skip$k$' refers to a skip connection with layer $k$. `resblock' denotes a residual block consisting of conv, instance norm, relu, conv, instance norm.
\subsubsection{Neural renderer}
The neural renderer architecture is a modified pix2pix network~\cite{pix2pix}:
\begin{center}
\begin{tabular}{ |c|c|c|c|c| } 
 \hline
 & layer type(s) & out channels & stride & activation\\
 \hline
 1 & $4\times 4$ conv & 64 & 2 & leaky\\
 2 & $4\times 4$ conv, bn & 128 & 2 & leaky\\
 3 & $4\times 4$ conv, bn & 256 & 2 & leaky\\
 4 & $4\times 4$ conv, bn & 256 & 2 & leaky\\
 5 & $4\times 4$ conv, bn & 256 & 2 & leaky\\
 6 & $4\times 4$ conv, bn & 256 & 1 & leaky\\
 7 & $4\times 4$ conv, bn & 256 & 1 & leaky\\
 8 & skip5, $4\times 4$ convt, bn & 256 & 2 & relu\\
 9 & skip4, $4\times 4$ convt, bn & 256 & 2 & relu\\
 10 & skip3, $4\times 4$ convt, bn & 128 & 2 & relu\\
 11 & skip2, $4\times 4$ convt, bn & 64 & 2 & relu\\
 12 & skip1, $4\times 4$ convt, bn & 64 & 2 & relu\\ 
 13 & $4\times 4$ conv & 4 & 1 & tanh\\
 \hline
\end{tabular}
\end{center}

\subsubsection{Upsampling network}
The upsampling network predicts a residual image for each bilinearly upsampled RGBA layer predicted by the neural renderer. The network input is the bilinearly upsampled (to the desired output size) concatenation of (1) the predicted low-resolution RGBA, (2) the sampled texture input, and (3) the final feature maps output by the neural renderer preceding the RGBA output layer. The RGBA outputs of the upsampling network are then composited according to $o_t$.
The upsampling network architecture is as follows:
\begin{center}
\begin{tabular}{ |c|c|c|c|c| } 
 \hline
 & layer type(s) & output channels & stride & activation\\
 \hline
 1 & $3\times 3$ conv, in & 64 & 1 & relu\\
 2 & $3\times 3$ resblock & 64 & 1 & relu\\
 3 & $3\times 3$ resblock & 64 & 1 & relu\\
 4 & $3\times 3$ resblock & 64 & 1 & relu\\
 5 & $3\times 3$ conv & 4 & 1 & none \\
 \hline
\end{tabular}
\end{center}

\subsubsection{Keypoints-to-UVs}
The keypoint-to-UV network is a fully convolutional network that takes in an RGB image of a skeleton and outputs a UV map of the same size. The architecture is the same as the neural renderer architecture, with the exception of the final layer, which is replaced by two heads: 1) a final convolutional layer with 25 output channels to predict body part and background classification, and 2) a convolutional layer with 48 output channels to regress UV coordinates for each of the 24 body parts. As in the DensePose work \cite{guler2018densepose}, we train the body part classifier with cross-entropy loss and train the predicted UV coordinates with L1 loss. The regression loss on the UV coordinates is only taken into account for a body part if the pixel lies within the specific part, as defined by the ground-truth UV map.

\end{document}